\documentclass[letterpaper]{article} 
\usepackage{aaai2026}  
\usepackage{times}  
\usepackage{helvet}  
\usepackage{courier}  
\usepackage[hyphens]{url}  
\usepackage{graphicx} 
\usepackage{natbib}  
\usepackage{caption} 
\usepackage{algorithm}
\usepackage{algorithmic}
\usepackage[colorlinks=true, 
            linkcolor=blue,       
            citecolor=green,      
           ]{hyperref}

\usepackage{newfloat}
\usepackage{listings}
\DeclareCaptionStyle{ruled}{labelfont=normalfont,labelsep=colon,strut=off} 
\floatstyle{ruled}
\newfloat{listing}{tb}{lst}{}
\floatname{listing}{Listing}
\usepackage{multirow}
\usepackage{booktabs}
\usepackage{amsmath} 
\usepackage{color} 

\title{MM-R5: MultiModal Reasoning-Enhanced ReRanker via Reinforcement Learning for Document Retrieval}

\author {
    Mingjun Xu\textsuperscript{\rm 1}\equalcontrib,
    Jinhan Dong\textsuperscript{\rm 1}\equalcontrib,
    Jue Hou\textsuperscript{\rm 1}\equalcontrib,
    Zehui Wang\textsuperscript{\rm 1},\\
    Sihang Li\textsuperscript{\rm 1},
    Zhifeng Gao\textsuperscript{\rm 1},
    Renxin Zhong\textsuperscript{\rm 2},
    Hengxing Cai\textsuperscript{\rm 1}\thanks{Corresponding authors.}
}
\affiliations {
    \textsuperscript{\rm 1}DP Technology, Beijing, China\\
    \textsuperscript{\rm 2}School of Intelligent Systems Engineering, Sun Yat-Sen University, Shenzhen, China\\
}

\usepackage{bibentry}

\begin{document}

\maketitle

\begin{abstract}
Multimodal document retrieval systems enable information access across text, images, and layouts, benefiting various domains like document-based question answering, report analysis, and interactive content summarization. Rerankers improve retrieval precision by reordering retrieved candidates.  However, current multimodal reranking methods remain underexplored, with significant room for improvement in both training strategies and overall effectiveness. Moreover, the lack of explicit reasoning makes it difficult to analyze and optimize these methods further. In this paper, We propose MM-R5, a \textbf{M}ulti\textbf{M}odal \textbf{R}easoning-Enhanced \textbf{R}e\textbf{R}anker via \textbf{R}einforcement Learning for Document \textbf{R}etrieval, aiming to provide a more effective and reliable solution for multimodal reranking tasks. MM-R5 is trained in two stages: supervised fine-tuning (SFT) and reinforcement learning (RL). In the SFT stage, we focus on improving instruction-following and guiding the model to generate complete and high-quality reasoning chains. To support this, we introduce a novel data construction strategy that produces rich, high-quality reasoning data. In the RL stage, we design a task-specific reward framework, including a reranking reward tailored for multimodal candidates and a composite template-based reward to further refine reasoning quality. We conduct extensive experiments on MMDocIR, a challenging public benchmark spanning multiple domains. MM-R5 achieves state-of-the-art performance on most metrics and delivers comparable results to much larger models on the remaining ones. Moreover, compared to the best retrieval-only method, MM-R5 improves recall@1 by over 4\%. These results validate the effectiveness of our reasoning-enhanced training pipeline. Our code is available at \url{https://github.com/i2vec/MM-R5}.
\end{abstract}


\section{Introduction}
\begin{figure}[t]
\centering
\includegraphics[width=0.47\textwidth]{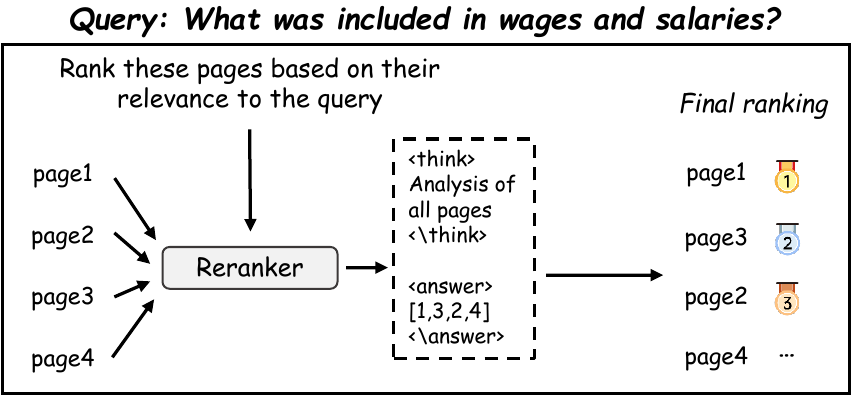} 
\caption{Workflow of MM-R5. It takes all candidate pages at once, analyzes all images, and then outputs the reasoning process along with a relevance ranking.}
\label{figure:intro}
\end{figure}
Multimodal document retrieval enhances traditional retrieval by incorporating visual modalities,~\cite{cui2021document, 10322990, lee2024unified} offering substantial benefits for tasks such as document-based question answering, report analysis, and interactive content summarization~\cite{hudson2019gqa, marino2019ok}. The quality of retrieval results serves as a critical determinant of the overall performance in downstream tasks~\cite{nogueira2019passage, zhuang2025rankr1enhancingreasoningllmbased, moreira2024enhancing}, and rerankers further enhance retrieval quality by performing fine-grained sorting of all candidate samples prior to generation. In text-only settings, rerankers leveraging large language models have demonstrated robust performance in document ranking tasks~\cite{ma2023zero, sun2025investigation, xu2024rankmamba, zhuang2023open, sun2023chatgpt}. Nevertheless, adapting these approaches to multimodal contexts presents substantial challenges due to the added complexity introduced by visual content. 

\indent Recent advances in vision-language models (VLMs) have shown promising potential in multimodal reranking tasks. However, despite some exploratory efforts in this field~\cite{chen2024mllm, MonoQwen}, significant limitations still remain: (i) current training-based approaches are limited both in number and effectiveness, although general-purpose VLMs can be used as rerankers, their performance in instruction-following and task-specific reasoning is often suboptimal without fine-tuning, leaving substantial room for performance improvement and methodological exploration; (ii) to the best of our knowledge, no existing method in multimodal reranking has incorporated Chain-of-Thought (CoT) reasoning. This absence results in a lack of interpretability in how the model assesses the relevance between candidates and the query, thereby hindering further analysis and optimization. To address these gaps, we aim to develop a more powerful multimodal reranker that explicitly integrates CoT, enabling both improved performance and interpretable reasoning.

\indent To this end, we design a two-stage training framework consisting of supervised fine-tuning (SFT) and reinforcement learning (RL). In the first stage, we construct a dataset containing high-quality reasoning paths and apply SFT to help the model adapt to multi-page reasoning and develop structured reasoning patterns. In the second stage of RL, inspired by prior work~\cite{guo2025deepseek, zhuang2025rankr1enhancingreasoningllmbased}, we leverage Group Relative Policy Optimization (GRPO) to further enhance the model’s reasoning ability, enhancing reranking performance while maintaining logical clarity and structural consistency in the reasoning process. Based on this design, we obtain MM-R5, a reasoning-enhanced multimodal reranker designed to overcome the limitations of existing methods. Built upon VLMs and empowered by our two-stage training scheme, MM-R5 demonstrates strong multi-page understanding capabilities and produces interpretable, step-by-step reasoning during the reranking process. As illustrated in Figure~\ref{figure:intro}, our model jointly process all candidate pages, analyze their relevance to the query individually, and produce both an explicit reasoning chain and a final ranking. This approach enabled us to achieve second place in the WWW 2025 Multimodal RAG Challenge. Our main contributions are as follows:
\begin{itemize}
\item We design a two-stage training pipeline to create MM-R5, a powerful reranker specifically for multimodal document retrieval. Specifically, we construct a high-quality dataset with well-annotated reasoning paths for supervised fine-tuning, and combine it with a task-specific reinforcement learning phase using GRPO to develop MM-R5.
\item We conduct experiments on publicly available benchmark datasets, where our method achieves state-of-the-art (SoTA) performance across multiple evaluation metrics, while remaining metrics are comparable or even better to those of significantly larger models.
\item We integrate MM-R5 with various retrieval models and consistently observe performance improvements, demonstrating the stability, effectiveness, and strong adaptability of our approach.
\item Our method generates explicit reasoning chains, enhancing interpretability and offering a new path toward more controllable and trustworthy multimodal retrieval systems.
\end{itemize}

\section{Related Work}
\subsection{Multimodal Document Retrieval}
Multimodal document retrieval focuses on extracting relevant information from documents that incorporate rich visual structures, including images, tables, and complex layouts. Driven by the rapid advancement of VLMs~\cite{alayrac2022flamingo, bai2025qwen2}, retrieval methodologies have transitioned from conventional text-only paradigms to multimodal approaches that integrate textual and visual data. Recent investigations~\cite{wei2024uniir, koukounas2024jina} delve into retrieval across diverse combinations of textual and visual inputs, striving to bridge the modality gap and foster more holistic retrieval frameworks. DSE~\cite{ma2024unifying} leverages large VLMs to encode document screenshots directly into dense representations, facilitating efficient retrieval without OCR dependency. ColPali~\cite{faysse2024colpali} enhances performance further by constructing high-quality multi-vector embeddings derived from document images and implementing a late interaction mechanism, achieving notable outcomes.

\indent Despite these advantages, existing methods mainly focus on retrieval performance, with the exploration of multimodal rerankers remaining limited. Therefore, in this paper, we aim to contribute an effective approach specifically for multimodal reranking.

\begin{figure*}[t]
\centering
\includegraphics[width=\textwidth]{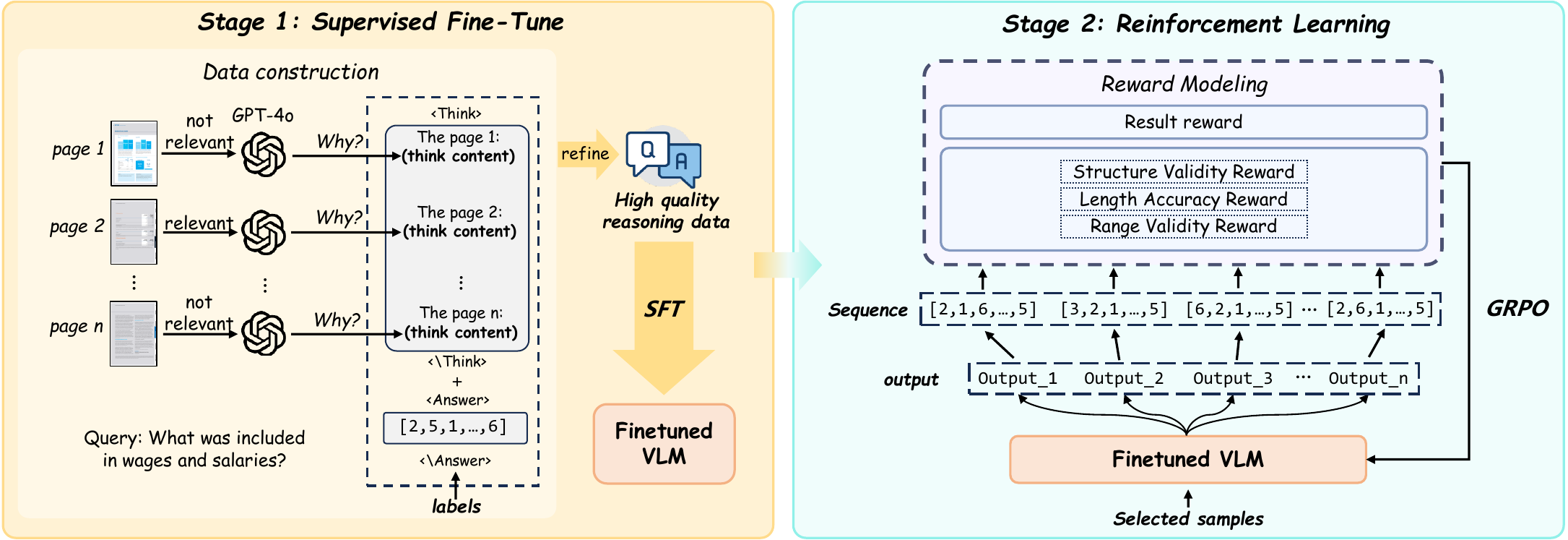} 
\caption{Overview of our two-stage training pipeline.}
\label{fig: overall}
\end{figure*}

\subsection{Reranker}
The reranker plays a pivotal role in the fine-grained reordering of candidates retrieved by the initial retriever, thereby enhancing the alignment between results and the user’s query. For RAG systems, effective high-quality reranking not only significantly boosts reasoning efficiency but also minimizes the risk of introducing noise. In recent years, large language models (LLMs) have been widely adopted for reranking tasks owing to their superior capabilities in language understanding and generation~\cite{zhuang2023beyond, guo2024generating, luo2024prp}. Likewise, VLMs have begun preliminary explorations into multimodal reranking~\cite{chen2024mllm, MonoQwen}, although current research in this domain is still nascent compared to unimodal scenarios. The complexity of multimodal data amplifies the noise caused by incorrect retrievals, thereby escalating the importance of retrieval accuracy in multimodal RAG systems. Designing a more efficient multimodal reranker has thus become the key to overcoming the performance bottleneck of current multimodal RAG frameworks.

\subsection{Multimodal Reasoning}
Recent concurrent studies have achieved substantial advancements in multimodal reasoning. For example, R1-OneVision~\cite{yang2025r1} developed a cross-modal reasoning framework that converts images into structured visual representations. These representations are subsequently utilized to construct a visual reasoning dataset with the assistance of a language model.  The Vision-Language Model (VLM) is initially trained on this dataset and further refined through RL.  Similarly, R1-V~\cite{chen2025r1v} integrates the GRPO method~\cite{shao2024deepseekmath} Reasoning in Open Language Models] from DeepSeek R1~\cite{guo2025deepseek} into VLM training, achieving remarkable performance in object-counting tasks—allowing a 3B model to outperform its 72B counterpart.

\indent Vision-R1~\cite{huang2025vision} adopts a data-centric approach by generating a multimodal CoT dataset derived from visual content. This dataset is used for cold-start training and subsequently enhanced via GRPO. VisualThinker-R1-Zero~\cite{zhou2025r1} demonstrated that applying R1 techniques to a base VLM rather than an instruction-tuned model yields greater performance improvements and induces the so-called "visual aha moment." A similar phenomenon was observed in MM-Eureka~\cite{meng2025mm}, which employed RLOO~\cite{ahmadian2024back} to train both an 8B instruction-tuned VLM and a 38B base model, resulting in more sophisticated and extended reasoning outputs.

\indent Curr-ReFT~\cite{deng2025boosting} proposed a three-stage RL framework featuring progressively challenging reward levels, while LMM-R1~\cite{peng2025lmm} introduced a rule-based two-stage RL strategy that first enhances text-based reasoning before addressing complex multimodal tasks. Although many of these works focus on multimodal mathematical reasoning~\cite{lu2023mathvista, wang2024measuring, zhang2024mathverse}, studies such as Visual-RFT~\cite{liu2025visual} and VLM-R1~\cite{shen2025vlm} extend RL-based reasoning to visual perception tasks.

\indent Building upon these insights, we observe that reasoning has consistently demonstrated its effectiveness across a wide range of multimodal tasks, from visual question answering to complex mathematical reasoning. These advances naturally lead us to explore the incorporation of reasoning into multimodal reranking.

\begin{figure*}[t]
\centering
\includegraphics[width=\textwidth]{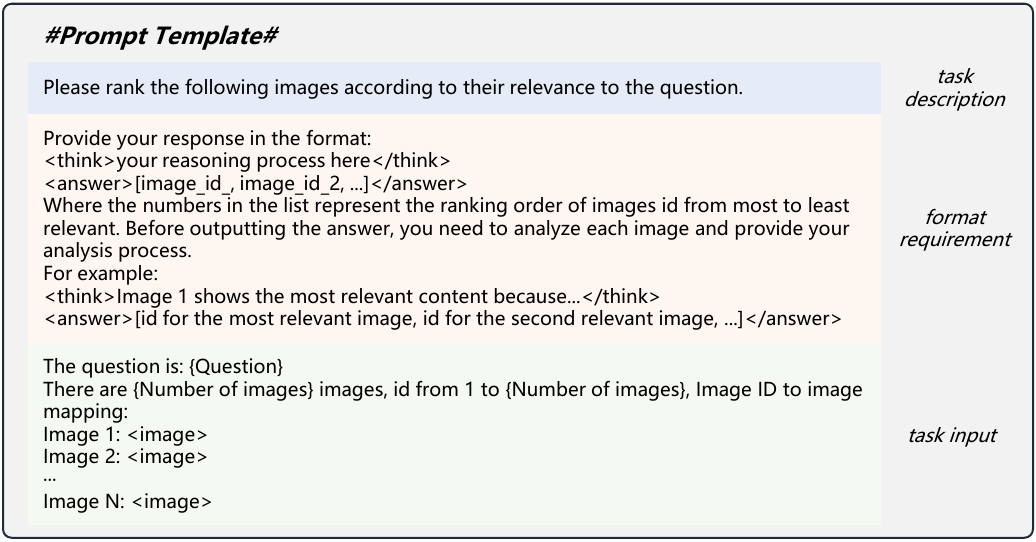} 
\caption{Prompt template for Chain-of-Thought based Reranking.}
\label{fig: prompt}
\end{figure*}
\section{Method}
In this section, we provide a detailed description of MM-R5. Specifically, we begin with a clear definition of the task, then we describe our two-stage training process, including the data construction and training procedure for each stage. However, our focus differs slightly between the two: in the first stage of SFT, we emphasize the construction of high-quality reasoning data; while in the second stage of reinforcement learning, we focus on the design of the reward function and its significance.
\subsection{Problem Formulation}
We address the task of multimodal reranking, which aims to reorder a set of retrieved multimodal candidates based on their relevance to a user query. The objective is to produce a ranking that prioritizes the most relevant items to support downstream generation tasks.

\indent Formally, each task instance consists of:
\begin{itemize}
  \item $q$, a natural language query representing the user's information need;
  \item $\mathcal{D} = \{d_1, d_2, \ldots, d_n\}$, a set of retrieved multimodal candidates, where each $d_i$ contains visual and textual information.
\end{itemize}

\indent Given the query $q$ and the candidate set $\mathcal{D}$, the reranker outputs a predicted index list $\hat{\mathcal{I}}$ that specifies a ranking permutation over $\mathcal{D}$, where smaller indices indicate higher predicted relevance to the query $q$. For evaluation, we denote the golden set of relevant candidate indices as $\mathcal{G}$, where $|\mathcal{G}|\le| \mathcal{D}|$.

\subsection{Stage 1: Supervised Fine-Tune}
\label{subsec:sft}
\subsubsection{Data Construction}
For general reasoning tasks, a common data construction pipeline involves prompting a strong language model to generate multiple responses containing reasoning paths based on the original question, followed by rule-based or manual rejection sampling to select high-quality samples. However, this approach presents significant limitations in the context of multimodal reranking, as addressed in this work. First, existing VLMs often struggle with multi-image inputs, especially when the number of images increases, resulting in poor reasoning quality. Second, in relevance ranking tasks, these models tend to over-focus on the most obviously relevant candidates while neglecting a thorough analysis of the less relevant ones, ultimately leading to suboptimal ranking accuracy.

\indent To address these issues and construct high-quality SFT data with explicit reasoning paths, we design a novel data construction pipeline based on GPT-4o, as illustrated in Figure~\ref{fig: overall}. Unlike traditional methods that directly ask the model to reason over multiple images simultaneously, our pipeline adopts a single-image reasoning strategy. Specifically, we pair each image with the original question to create multiple single-image sub-tasks. For each sub-task, the model is instructed to explain why the image is or is not relevant to the query, guided by its label, resulting in individual image-level reasoning statements. These statements are then matched with their corresponding image indices and concatenated in order to form a complete reasoning chain, which is placed within \texttt{<think>...</think>} tags to serve as the model’s reasoning path. And for the final ranked list of image indices, we sort the images based on their relevance labels—placing relevant images at the front and irrelevant ones at the end—and wrap the list in \texttt{<answer>...</answer>} tags. By combining this structured \texttt{<think>...</think> <answer>...</answer>} response with the original prompt, we obtain a high-quality SFT training sample. We then leverage GPT-4o to further refine the reasoning path of this sample, making it more concise while preserving the completeness and clarity.

\indent Using this pipeline, we constructed 7,200 high-quality training instances for SFT, which played a crucial role in enhancing the model’s instruction-following and reasoning capabilities.

\subsubsection{Training}
After obtaining a dataset with high-quality reasoning paths, we fine-tune a pretrained VLM on it. This stage enables the model to better align with the multimodal reranking task by significantly enhancing its multi-page reasoning abilities and fostering systematic reasoning patterns. In particular, the model learns to explicitly compare each retrieved image with the query, evaluating their relevance individually rather than relying on shallow or incomplete reasoning. This mitigates the risk of missing relevant candidates due to under-reasoning, a common failure mode in general-purpose VLMs. As a result, this supervised fine-tuning phase lays a solid foundation for the subsequent reinforcement learning stage, where the model’s reranking ability is further optimized.

\subsection{Stage 2: Reinforcement Learning}
\label{subsec:rl}
\subsubsection{Resolution-Balanced Sampling Strategy}
Unlike the SFT stage, which requires constructing task-specific data, the training data for the reinforcement learning phase is directly sampled from the original training corpus.

\indent To mitigate the potential bias introduced by image resolution variations during reinforcement learning, we design a resolution-balanced data sampling strategy. Specifically, the original training dataset is first partitioned into 10 subsets, where each subset contains samples with approximately similar image sizes. To ensure uniform representation across different resolution ranges, we perform sampling without replacement from each subset. From the full dataset, we select a total of 3,000 samples, allocating them proportionally across all subsets to form the final resolution-balanced  training set.

\indent Since image resolutions within each subset are relatively consistent, this proportional sampling ensures that all resolution segments contribute equally during training, without increasing the total data volume. This strategy effectively reduces gradient estimation bias caused by imbalanced resolution distributions, leading to more stable and reliable reinforcement learning.

\subsubsection{Reward Modeling}
As illustrated in Fig~\ref{fig: overall}, after obtaining the resolution-balanced training data, we perform reinforcement learning on the model using GRPO. To enhance the model's reranking capability while ensuring that its output remains precise and well-formatted, we design two complementary reward functions: a result reward and a format reward.


 \noindent\textbf{Result reward.} We propose a novel reward formulation specifically designed to optimize relevance-based reranking performance. Let the candidate set be denoted by $\mathcal{D} = \{d_1, d_2, \dots, d_n\}$, and let $\hat{\mathcal{I}} = [\hat{i}_1, \hat{i}_2, \dots, \hat{i}_k]$ represent the predicted indices over $\mathcal{D}$ output by the reranker. The golden set of relevant item indices is denoted as $G \subseteq \{1, \dots, n\}$. For each predicted index $\hat{i}_j \in \hat{\mathcal{I}}$, we define a binary indicator $s_j$, where $s_j = 1$ if $\hat{i}_j \in G$, and $s_j = 0$ otherwise.


\begin{align}
R_{result} = \frac{\displaystyle\sum_{j=1}^{|\hat{\mathcal{I}}|} \frac{s_j}{j^{3}}}{\displaystyle\sum_{j=1}^{|\mathcal{G}|} \frac{1}{j^{3}}}, \quad s_j =
\begin{cases}
1, & \text{if } \hat{i}_j \in \mathcal{G} \\
0, & \text{if } \hat{i}_j \notin \mathcal{G}
\end{cases} 
\end{align}

\indent The numerator sums the inverse cubes of the predicted ranks for all correctly identified items, which gives much higher weight to items ranked at the top, especially the first position. The denominator represents the maximum possible reward when all ground-truth items are ranked in the top $K$ positions. This normalization ensures the final reward $R$ falls within the range $[0, 1]$ and reduces the effect of varying numbers of ground-truth items across samples. Compared to standard NDCG with logarithmic discounting, using the cube of the rank provides stronger emphasis on top-ranked predictions.


\noindent\textbf{Format reward.} In addition to relevance-oriented optimization, we define a format reward $R_{\text{format}}$ to ensure that the model produces structurally correct outputs. This reward is composed of three independent and differentiable components that jointly evaluate the legality of output formatting, the deviation in list length, and the validity of index values. Let $o$ denote the raw model output, and let $\hat{\mathcal{I}}$ denote the predicted index list derived from $o$ after parsing and deduplication. Let $G \subseteq \{1, \dots, n\}$ be the golden set of relevant indices, where $N$ is the total number of reference candidates. The list $\hat{\mathcal{I}}$ is treated as a set-valued list, in which any duplicate indices are merged to ensure uniqueness. The reward components are defined as follows:

\begin{itemize}
    \item \textbf{Structure Validity Reward}: 
    If the output sequence $o$ is properly enclosed within the required tags \texttt{<think>}...\texttt{</think>} and \texttt{<answer>}...\texttt{</answer>}, we assign a structure validity reward $R_{\text{valid}} = 1$; otherwise, $R_{\text{valid}} = 0$. This hard constraint ensures that any formatting error leads to a zero format reward, regardless of content correctness.
    
    \item \textbf{Length Accuracy Reward}: 
    Let $abs(\cdot)$ donate the absolute value function. The length accuracy reward is defined as:
    \begin{align}
    R_{\text{len}} = 1 - \frac{abs(|\hat{\mathcal{I}}| - n)}{n}
    \end{align}
    This component encourages the model to generate answers of equal length to the reference.

    \item \textbf{Range Validity Reward}:
    To ensure that predicted indices remain within the valid range, we define the reward as follows:
    \begin{align}
        R_{\text{range}} = \frac{ \sum_{i \in \hat{\mathcal{I}}} \delta(i)}{|\hat{\mathcal{I}}|}, \quad 
        \delta(i) =
        \begin{cases}
            1, & \text{if } 1 \le i \le n, \\
            0, & \text{otherwise}.
        \end{cases}
    \end{align}
    
    This reward encourages the model to generate indices strictly within the valid range and penalizes any out-of-bound predictions.
\end{itemize}

\indent The final format reward is computed as:
\begin{align}
R_{\text{format}} = R_{\text{valid}} \times R_{\text{len}} \times R_{\text{range}}.
\end{align}
\indent This formulation provides the following benefits: (i) $R_{\text{valid}}$ enforces strict adherence to the output structure through a hard-stop penalty, (ii) $R_{\text{len}}$ and $R_{\text{range}}$ offer smooth gradient signals when outputs are nearly correct, (iii) the three factors are independently adjustable, allowing fine-grained control over the penalty strength for different error types during training.

\subsubsection{Prompt Design}
To ensure fairness and consistency across all CoT based rerankers, we adopt a unified prompting strategy that standardizes the reasoning process by explicitly separating reasoning and final answers using structured markers, as illustrated in Figure~\ref{fig: prompt}. The prompt consists of three components: (i) a target description, which instructs the model to rank images based on their relevance to the query, (ii) a format requirement, which clearly defines the expected output structure, and (iii) a structured task input containing the query and a list of images, notably, each image is preceded by a textual identifier indicating its image ID, which helps the model better associate the visual content with its corresponding ID. This carefully designed prompt not only enforces output consistency across models but also improves interpretability and grounding by explicitly linking image content with their IDs.

\section{Experiments}
\begin{table*}[t]
\centering
\belowrulesep=-1pt
\aboverulesep=0pt
\renewcommand{\arraystretch}{1.4}\resizebox{\textwidth}{!}{
\begin{tabular}{c|lcccccc}
\hline
\multirow{2}{*}{Algorithm}               & \multicolumn{1}{c}{\multirow{2}{*}{Method}} & \multicolumn{3}{c}{Macro}                                                          & \multicolumn{3}{c}{Micro}                                                          \\
\cmidrule(lr){3-5} \cmidrule(l){6-8}
                                         & \multicolumn{1}{c}{}                        & \multicolumn{1}{c}{Recall@1} & \multicolumn{1}{c}{Recall@3} & \multicolumn{1}{c}{Recall@5} & \multicolumn{1}{c}{Recall@1} & \multicolumn{1}{c}{Recall@3} & \multicolumn{1}{c}{Recall@5} \\ \hline
\multirow{5}{*}{Retriever-only}          & CLIP~\cite{radford2021learningclip}                                        & 0.3334                       & 0.5428                       & 0.6452                       & 0.3227                       & 0.5337                       & 0.6350                       \\
                                         & E5-V~\cite{jiang2024e5v}                                        & 0.4201                       & 0.6470                       & 0.7389                       & 0.4249                       & 0.6578                       & 0.7519                       \\
                                         & DSE~\cite{ma2024unifying}                                         & 0.5109                       & 0.7194                       & 0.7925                       & 0.5023                       & 0.7234                       & 0.7971                       \\
                                         & GME~\cite{zhang2024gme}                                         & 0.5400                       & 0.7603                       & 0.8308                       & 0.5421                       & 0.7603                       & 0.8377                       \\
                                         & ColQwen~\cite{faysse2024colpali}                                     & 0.6481                       & 0.8331                       & 0.8766                       & 0.6354                       & 0.8213                       & 0.8667                       \\ \hline
\multirow{10}{*}{Reranking from ColQwen} & RagVL                                       & 0.3814                       & 0.6462                       & 0.7667                       & 0.3411                       & 0.6206                       & 0.7497                       \\
                                         & Qwen2.5-VL-7B~\cite{bai2025qwen2}                               & 0.6230                       & 0.8243                       & 0.8675                       & 0.6012                       & 0.8069                       & 0.8555                       \\
                                         & Qwen2.5-VL-7B-cot                           & 0.6479                       & 0.8179                       & 0.8670                       & 0.6336                       & 0.8103                       & 0.8604                       \\
                                         & Qwen2.5-VL-32B~\cite{bai2025qwen2}                              & 0.6422                       & 0.8161                       & 0.8669                       & 0.6309                       & 0.8118                       & 0.8636                       \\
                                         & Qwen2.5-VL-32B-cot                          & \underline{0.6768}                       & \underline{0.8500}                       & \underline{0.8842}                       & \underline{0.6609}                       & \textbf{0.8448}                       & \textbf{0.8800}                       \\
                                         & Gemma3-12B~\cite{team2025gemma}                                      & 0.4743                         & 0.6675                         & 0.7278                         & 0.4456                         & 0.6353                         & 0.7025                         \\
                                         & Gemma3-12B-cot                                  & 0.5403                         & 0.7729                         & 0.8292                         & 0.5166                         & 0.7536                         & 0.8153                         \\
                                         & Qwen2.5-VL-7B-sft                                     & 0.6673                       & 0.8475                       & 0.8828                       & 0.6498                       & 0.8366                       & 0.8746                       \\
                                         & Qwen2.5-VL-7B-rl                                      & 0.6586                       & 0.8370                       & 0.8827                       & 0.6454                       & 0.8309                       & 0.8744                       \\
                                         & MM-R5(ours)                                    & \textbf{0.6951}              & \textbf{0.8520}              & \textbf{0.8842}              & \textbf{0.6759}              & \underline{0.8401}              & \underline{0.8755}             \\ \hline
\end{tabular}
}
\caption{Performance on MMDocIR. ``Retriever-only'' indicates that we only used the results from the Retriever as the final output, and ``Reranking from ColQwen'' means that we used the reranker model to reorder the candidates output by ColQwen. Methods suffixed with ``cot'' require the model to generate a chain-of-thought reasoning process. All three of our methods incorporate reasoning chains. “MM-R5” refers to the model trained with the full two-stage pipeline, while the variants marked with “sft” and “rl” indicate models trained only with supervised fine-tuning or reinforcement learning, respectively. The bolded metrics indicate the best performance, while the underlined metrics represent the second best.}
\label{tab: main result}
\end{table*}
\subsection{Experimental Settings}
\subsubsection{Dataset}
We conduct our experiments on MMDocIR~\cite{dong2025mmdocir}, a large-scale benchmark specifically designed for multimodal document retrieval and reasoning. MMDocIR comprises both a training set and an evaluation set, providing rich modality diversity and reflecting the complexity of real-world documents.\\
\indent The training set includes 6,878 documents and 73,843 QA pairs sourced from seven existing DocVQA-related datasets, e.g., MP-DocVQA~\cite{tito2023hierarchical}, SlideVQA~\cite{tanaka2023slidevqa}, TAT-DQA~\cite{zhu2022towards}, SciQAG~\cite{wan2024sciqag}, DUDE~\cite{van2023document}, CUAD~\cite{hendrycks2021cuad}, spanning diverse domains and document lengths. Meanwhile, the evaluation set comprises 313 lengthy documents, averaging 65.1 pages each, spanning ten distinct domains such as academic, legal, and financial texts. Each domain is treated as an individual subset for evaluation. These documents integrate multiple modalities, including text, images, tables, and layout/meta elements, and are accompanied by 1,658 expert-annotated questions. Each question is annotated with both page-level and layout-level ground-truth labels, enabling fine-grained and precise assessment. Importantly, many questions demand complex reasoning that involves cross-modal understanding, multi-page synthesis, and layout-aware interpretation, making this benchmark particularly challenging.\\
\indent We adopt MMDocIR for both training and evaluation due to its comprehensive supervision signals and its ability to assess a model's capacity to retrieve and reason over complex multimodal content at the page level.

\begin{table*}[t]
\centering
\renewcommand{\arraystretch}{1.3}\resizebox{\textwidth}{!}{

\begin{tabular}{lcllllll}
\hline
\multicolumn{1}{c}{\multirow{2}{*}{Retriever}}          & \multirow{2}{*}{MM-R5} & \multicolumn{3}{c}{Macro}                                                                                                    & \multicolumn{3}{c}{Micro}                                                                                             \\
\cmidrule(lr){3-5} \cmidrule(l){6-8}
\multicolumn{1}{c}{}                                    &                        & \multicolumn{1}{c}{Recall@1}             & \multicolumn{1}{c}{Recall@3}              & \multicolumn{1}{c}{Recall@5}          & \multicolumn{1}{c}{Recall@1}          & \multicolumn{1}{c}{Recall@3}          & \multicolumn{1}{c}{Recall@5}          \\ \hline
\multirow{2}{*}{CLIP~\cite{radford2021learningclip}}    & w/o                    & 0.3334                                   & 0.5428                                    & 0.6452                                & 0.3227                                & 0.5337                                & 0.6350                                \\
                                                        & w                      & 0.5942 (\textcolor{green}{+26.08\%})     & 0.6749 (\textcolor{green}{+13.21\%})      & 0.7173 (\textcolor{green}{+7.21\%})   & 0.5803 (\textcolor{green}{+25.76\%})  & 0.6598 (\textcolor{green}{+12.61\%})  & 0.7018 (\textcolor{green}{+6.68\%})   \\ \hline
\multirow{2}{*}{E5-V~\cite{jiang2024e5v}}               & w/o                    & 0.4201                                   & 0.6470                                    & 0.7389                                & 0.4249                                & 0.6578                                & 0.7519                                \\
                                                        & w                      & 0.6236 (\textcolor{green}{+20.35\%})     & 0.7508 (\textcolor{green}{+10.38\%})      & 0.7966 (\textcolor{green}{+5.77\%})   & 0.6321 (\textcolor{green}{+20.72\%})  & 0.7516 (\textcolor{green}{+9.38\%})   & 0.8009 (\textcolor{green}{+4.90\%})   \\ \hline
\multirow{2}{*}{DSE~\cite{ma2024unifying}}              & w/o                    & 0.5109                                   & 0.7194                                    & 0.7925                                & 0.5023                                & 0.7234                                & 0.7971                                \\
                                                        & w                      & 0.6487 (\textcolor{green}{+13.78\%})     & 0.7777 (\textcolor{green}{+5.83\%})       & 0.8213 (\textcolor{green}{+2.88\%})   & 0.6429 (\textcolor{green}{+14.06\%})  & 0.7817 (\textcolor{green}{+5.83\%})   & 0.8254 (\textcolor{green}{+2.83\%})   \\ \hline
\multirow{2}{*}{GME~\cite{zhang2024gme}}                & w/o                    & 0.5400                                   & 0.7603                                    & 0.8308                                & 0.5421                                & 0.7603                                & 0.8377                                \\
                                                        & w                      & 0.6612 (\textcolor{green}{+12.12\%})     & 0.8095 (\textcolor{green}{+4.92\%})       & 0.8578 (\textcolor{green}{+2.70\%})   & 0.6551 (\textcolor{green}{+11.30\%})  & 0.8037 (\textcolor{green}{+4.34\%})   & 0.8566 (\textcolor{green}{+1.89\%})   \\ \hline
\multirow{2}{*}{ColQwen~\cite{faysse2024colpali}}       & w/o                    & 0.6481                                   & 0.8331                                    & 0.8766                                & 0.6354                                & 0.8213                                & 0.8667                                \\
                                                        & w                      & 0.6951 (\textcolor{green}{+4.70\%})      & 0.8520 (\textcolor{green}{+1.89\%})       & 0.8842 (\textcolor{green}{+0.76\%})   & 0.6759 (\textcolor{green}{+4.05\%})   & 0.8401 (\textcolor{green}{+1.88\%})   & 0.8755 (\textcolor{green}{+0.88\%})   \\ \hline
\end{tabular}
}
\caption{Extended experiments validating the effectiveness of MM-R5 across different retrievers. ``w" means reranking is applied using MM-R5, while ``w/o" refers to using original retrieval results without reranking.}
\label{tab: extend result}
\end{table*}

\subsubsection{Evaluation Metrics}
We follow~\cite{dong2025mmdocir} and adopt Recall@$k$ as the primary evaluation metric. MM-R5 scores each page in the document based on its relevance to the input question and returns the top-$k$ pages with the highest scores. Recall@$k$ is defined as the proportion of ground-truth pages that appear within the top-$k$ retrieved results. This metric reflects the model's ability to correctly identify and prioritize the most relevant pages containing evidence required to answer the question. In addition, We report both micro and macro Recall@k: micro averages recall across all individual samples, while macro first computes the average recall within each subset and then averages across subsets, providing a more balanced view across varying data distributions.


\subsubsection{Model and Training Configuration}
We adopt Qwen2.5-VL-7B~\cite{bai2025qwen2} as the baseline model for our two-stage training. In the first stage, we fine-tune the baseline model using the swift~\cite{zhao2024swiftascalablelightweightinfrastructure}, which allows efficient SFT on our carefully constructed dataset. This process yields a finetuned model with enhanced instruction-following and reasoning capabilities. In the second stage, we further train the model using the GRPO algorithm within the open-source VLM-R1~\cite{shen2025vlm} framework, where Low-Rank Adaptation (LoRA) is applied to improve training efficiency.

\indent The training is conducted with the following hyperparameter settings:
\begin{itemize}
    \item In the stage of supervised fine-tune, the training runs for 1 epoch with a learning rate of $1 \times 10^{-4}$, a batch size of 1, and gradient accumulation over 4 steps. LoRA is applied for efficient parameter tuning, with the rank lora\_r set to 8 and the scaling factor lora\_alpha set to 32.
    \item In the stage of reinforcement learning, the rollout number is set to 4, and the training runs for 1 epoch. We use a learning rate of $1 \times 10^{-5}$, a batch size of 1, and apply gradient accumulation over 2 steps. LoRA is used to improve parameter efficiency during fine-tuning, with the rank lora\_r set to 64 and the scaling factor lora\_alpha set to 128. 
\end{itemize}
All experiments are carried out on 4 NVIDIA A100 GPUs, each with 80GB of memory.
\subsubsection{Comparison Methods}
To demonstrate the effectiveness of our proposed method, we conduct a comprehensive evaluation across both retrieval and reranking settings. For retrieval baselines, we compare against several strong models, including DSE~\cite{ma2024unifying}, GME~\cite{zhang2024gme}, and ColQwen~\cite{faysse2024colpali}. While for reranking, we compare our method with the prior multimodal reranking model RAG-VL~\cite{chen2024mllm}, as well as representative vision-language models such as Qwen2.5-VL-32B~\cite{bai2025qwen2} and Gemma3-12B~\cite{team2025gemma}, covering a wide range of capabilities in vision-language understanding and instruction following. To ensure fair comparison in reranking, we fix the retrieval backbone to generate the top-10 candidate pages. This consistent candidate set allows us to isolate the impact of different reranking strategies under identical input conditions.



\begin{figure*}[t]
\centering
\includegraphics[width=\textwidth]{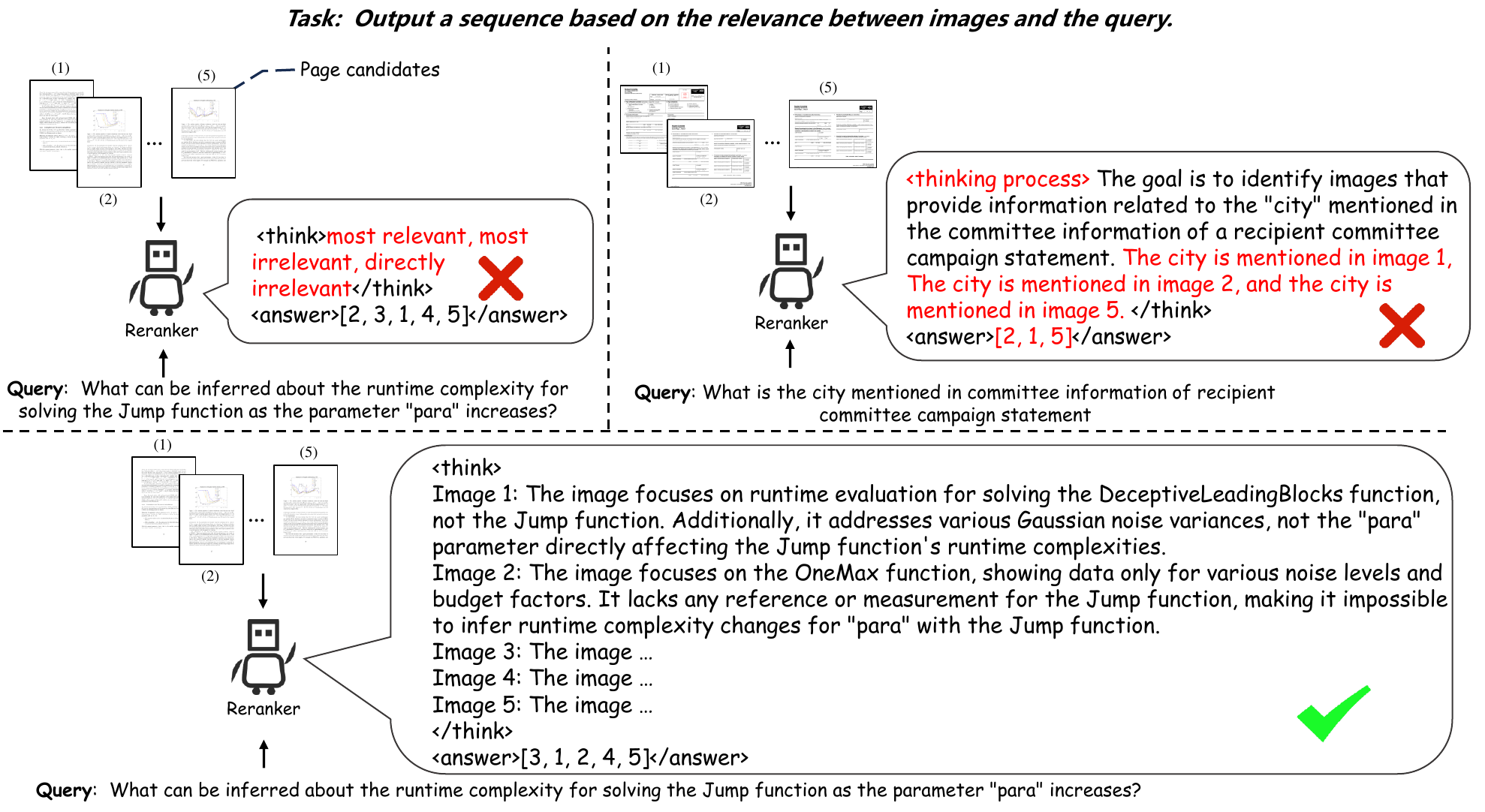} 
\caption{Comparison between our model and multimodal rerankers without training. All page candidates are retrieved by ColQwen. The red-highlighted parts in the model’s output indicate segments that are of poor quality. The example in the top-right illustrates that the reranker model not only fails to follow the instruction to produce a sequence as required but also exhibits insufficient reasoning in its response, while example in the top-left corner reveals that its reasoning process totally lacks interpretability. In contrast, our model demonstrates significantly improved instruction-following ability, and exhibits more coherent and interpretable reasoning steps.}
\label{fig:case}
\end{figure*}

\subsection{Experimental Results}
\subsubsection{Main Results} 
As illustrated in Table~\ref{tab: main result}, Our method, MM-R5, consistently achieves SoTA performance across most evaluation metrics, including both macro and micro recall at different thresholds. Compared with all retriever-only methods, MM-R5 brings significant improvements, for instance, compared with the previous SoTA retriever ColQwen, macro recall@1 improves from 0.6481 to 0.6951, and micro recall@1 improves from 0.6354 (ColQwen) to 0.6759, demonstrating the effectiveness of our reranking strategy in refining the initial retrieval results.

\indent Furthermore, when compared to strong reranking baselines, including large-scale general-purpose models such as Qwen2.5-VL-32B and Gemma3-12B, MM-R5 still achieves the best performance, despite being based on a much smaller baseline, Qwen2.5-VL-7B. Specifically, MM-R5 improves macro Recall@1 from 0.5403 to 0.6951 and micro Recall@1 from 0.5166 to 0.6759 compared to Gemma3-12B-cot, achieving remarkable gains of 0.1548 and 0.1593, respectively. It also surpasses the much larger Qwen2.5-VL-32B-cot by a notable margin in both metrics. These results collectively validate the design of our method: with effective supervision and well-designed training strategies, our reranker not only enhances retrieval performance but also outperforms much larger vision-language models.

\subsubsection{Ablation Studies}
In this part, we investigate the impact of incorporating CoT reasoning into multimodal reranking tasks, and examine the necessity of our proposed two-stage training framework.

\noindent \textbf{Impact of CoT.} As shown in Table~\ref{tab: main result}, incorporating CoT reasoning consistently improves reranking performance. For instance, Qwen2.5-VL-7B with CoT improves macro recall@1 by 2.49 and micro recall@1 by 3.24, while Qwen2.5-VL-32B achieves gains of 3.46 and 3.00 in macro and micro recall@1 respectively. These results suggest that CoT reasoning can effectively enhance the model’s reranking capability. In particular, our introduced reasoning paradigm, which guides the model to explicitly evaluate the relevance between each image and the query before producing the final ranking—not only leads to improved reranking performance, but also enhances the interpretability of the model's reasoning process. The discussion of the interpretability will be detailed in the case study section.

\noindent \textbf{Impact of two-stage training.} Table~\ref{tab: main result} also includes two ablated variants of our model: Qwen2.5-VL-7B-sft, trained only with supervised fine-tuning, and Qwen2.5-VL-7B-rl, trained only with reinforcement learning. Both fall short of the full model on every metric. For example, in macro Recall@1 the full model reaches 0.6951, whereas Qwen2.5-VL-7B-sft and Qwen2.5-VL-7B-rl obtain 0.6673 (-2.78) and 0.6586 (-3.65), respectively. A similar pattern appears in micro Recall@1: MM-R5 achieves 0.6759, surpassing the SFT-only and RL-only variants by 2.61 and 3.05. Although these single-stage variants substantially outperform the baseline Qwen2.5-VL-7B-cot (macro Recall@1: 0.6479, and micro Recall@1: 0.6336), the full two-stage model further amplifies these gains, confirming that SFT and RL are complementary, validating the necessity and synergy of our two-stage training pipeline.

\subsubsection{Generalization Across Retrievers}
To further evaluate the generalization ability of our reranker, we conduct extended experiments by applying MM-R5 on top of five representative multimodal retrievers: CLIP~\cite{radford2021learningclip}, E5-V~\cite{jiang2024e5v}, DSE, GME, and ColQwen. 

\indent As shown in Table~\ref{tab: extend result}, across all retrievers and evaluation metrics, applying MM-R5 consistently improves both macro and micro recall. For instance, on strong retrievers like GME, recall@1 improves from 0.5400 to 0.6612 (macro) and from 0.5421 to 0.6551 (micro); on ColQwen, recall@1 improves from 0.6481 to 0.6951 (macro) and from 0.6354 to 0.6759 (micro). In addition to recall@1, MM-R5 also brings consistent improvements on recall@3 and recall@5. These results demonstrate that MM-R5 is highly effective and robust across different retrieval backbones, confirming its general applicability as a plug-and-play reranking component. Importantly, the relative gains are particularly significant for weaker retrievers like CLIP and E5-V, indicating that MM-R5 not only refines strong baselines but also substantially enhances simpler retrievers.

\subsubsection{Case Study For Interpretability }
\label{subserc:case study}
In previous experiments, we demonstrated that incorporating CoT reasoning significantly boosts model performance. However, compared to MM-R5, directly prompting existing general untrained VLMs to produce CoT often leads to insufficient and unclear reasoning, resulting in poor interpretability. As illustrated in Figure~\ref{fig:case}, the upper examples use our baseline model, Qwen2.5-VL-7B, as the reranker, while the lower example shows results from our MM-R5. In the top-right case, the baseline model already struggles with format adherence, moreover, its reasoning merely notes that pages 1, 2, and 5 mention the word “city” from the query, failing to consider other relevant candidates and providing unclear, superficial explanations. The top-left example underscores the problem of vague and insufficient justifications, where the model provides no reasoning at all. In contrast, as shown in the lower part, when presented with the same sample as in the top-left example,  MM-R5 demonstrates strong adherence to the required output format and delivers well-structured analysis for each image before presenting the final answer. This highlights that our trained reranker is capable of generating interpretable CoT reasoning, which facilitates deeper analysis and supports future improvements.

\section{Conclusion}
In this paper, we propose MM-R5, a reasoning-enhanced multimodal reranker trained via a two-stage pipeline. In the first stage, supervised fine-tuning, we introduce a novel reasoning-oriented data construction strategy that provides high-quality instruction-following and reasoning examples to effectively guide the model’s initial learning. In the second stage, we design a reinforcement learning framework tailored to the multimodal reranking task, incorporating a customized reward function that promotes reranking effectiveness while preserving structured and well-formed outputs. Extensive experiments demonstrate the effectiveness of MM-R5 in improving reranking quality and highlight its generality across various retriever backbones.

\bibliography{aaai2026}

\end{document}